\documentclass[runningheads]{llncs}

\usepackage{eccv}

\usepackage{eccvabbrv}

\usepackage{graphicx}
\usepackage{booktabs}

\usepackage[accsupp]{axessibility}  %

\usepackage{hyperref}

\usepackage{orcidlink}
\usepackage{amsmath}
\DeclareMathOperator*{\argmax}{arg\,max}

\newcommand{\norm}[1]{\left\lVert#1\right\rVert}
\usepackage{colortbl}
\usepackage{multirow}
\definecolor{Gray}{gray}{0.9}
\definecolor{Gray0}{gray}{0.8}
\usepackage{ragged2e}
\usepackage{wrapfig}
\usepackage{algorithm}
\usepackage{caption}
\usepackage[skip=0.5ex, belowskip=1ex,
            labelformat=brace,
            singlelinecheck=off]{subcaption}
\usepackage{algorithmic} 
\usepackage{marvosym}

\usepackage{booktabs}
\usepackage{threeparttable}
\usepackage{makecell}
\usepackage{graphicx}
\newcommand{\DeepALPlus}{\textit{DeepAL$^{+}$ }}

\begin{document}

\title{Maximally Separated Active Learning}

\author{Tejaswi Kasarla\inst{1}\orcidlink{0000-0003-4580-9383} \and
Abhishek Jha\inst{2}\orcidlink{0000-0002-0350-2474} \and
Faye Tervoort\inst{1} \and
 Rita Cucchiara\inst{3,4}\orcidlink{0000-0002-2239-283X} \and Pascal Mettes\inst{1}\orcidlink{0000-0001-9275-5942}}
\authorrunning{Kasarla et al.}

\institute{University of Amsterdam, The Netherlands
\and 
KU Leuven, Belgium 
\and
University of Modena and Reggio Emilia, Italy 
\and
IIT-CNR, Italy }

\maketitle

\begin{abstract}

Active Learning aims to optimize performance while minimizing annotation costs by selecting the most informative samples from an unlabelled pool. Traditional uncertainty sampling often leads to sampling bias by choosing similar uncertain samples. We propose an active learning method that utilizes fixed equiangular hyperspherical points as class prototypes, ensuring consistent inter-class separation and robust feature representations. Our approach introduces Maximally Separated Active Learning (MSAL) for uncertainty sampling and a combined strategy (MSAL-D) for incorporating diversity. This method eliminates the need for costly clustering steps, while maintaining diversity through hyperspherical uniformity. We demonstrate strong performance over existing active learning techniques across five benchmark datasets, highlighting the method's effectiveness and integration ease. The code is available on github\addtocounter{footnote}{-4}\footnote{\url{https://github.com/tkasarla/MSAL}}.

  \keywords{Hyperspherical Learning \and Active Learning}
\end{abstract}

\section{Introduction}
The quality of training data has a significant impact on its performance, robustness, fairness and safety \cite{mohseni2022taxonomy}. Active learning is a popular technique for ensuring data quality \cite{li2024active} in safety critical applications such as autonomous driving \cite{kasarla2019region, xie2022towards}, medical imaging \cite{wang2024comprehensive}. 

With the abundant availability of large scale unlabeled datasets, it is often laborious and expensive to annotate for downstream applications. In practice, active learning (AL) aims to maximize a model's performance gain while annotating the fewest examples possible. This is achieved by strategically selecting the most informative unlabeled samples for human annotation.  Within AL alogrithms, uncertainty sampling is a commonly used method to determine the informativeness of data points \cite{Ash2020Deep,zhao2021uncertaintyaware,woo2023active} due to its simplicity and computational efficiency. This approach quantifies the uncertainty of a sample \cite{settles2009, settles2011theories} given a model, and selects unlabeled samples with greatest uncertainty for annotation. Various measures \cite{nguyen2022uncertainty} exist for determining the uncertainty of a sample.

However, relying on uncertainty sampling alone may not lead to optimal performance, as it can result in the selection of similar \textit{uncertain} samples, leading to \textit{sampling bias} \cite{DASGUPTA20111767}, a common challenge in uncertainty-based approaches. Several methods have been proposed to mitigate sampling bias by incorporating diversity into the selection process. These methods include submodular maximization \cite{wei2015submodular}, clustering\cite{zhdanov2019diverse,citovsky2021batch, yang2021batch}, k-means\texttt{++} seeding \cite{Ash2020Deep}, joint mutual information \cite{kirsch2019batchbald}. 

Learning good feature representations remains an open problem in active learning~\cite{zhan2022comparative}. The initial predictor model is often trained on a small labeled subset, which can result in the failure to learn robust discriminative features. This, subsequently, leads to underutilized metric space even when additional data is incorporated. Recent research in hyperspherical learning aims to address a similar issue in image classification tasks. Specifically, hyperspherical uniformity has proven effective for class separation, whether as additional objectives in a softmax cross-entropy optimization \cite{lin2020regularizing,liu2018learning,pmlr-v130-liu21d,zhou2022learning} or through hyperspherical prototypes \cite{mettes2019hyperspherical,Kasarla2022}. By ensuring hyperspherical uniformity of classes through fixed class prototypes\cite{Kasarla2022}, the representations become more stable and less prone to bias toward any particular class. This approach has also been shown to enhance classification performance on imbalanced datasets, thereby providing a strong foundation for its application in imbalanced active learning.

Our proposed method utilizes fixed equiangular hyperspherical points as class prototypes, introduced by Kasarla et.al. \cite{Kasarla2022}, for active learning in two ways. First, we incorporate these points to encode maximum separation as an inductive bias within the network architecture. This ensures consistent inter-class separation among class embeddings in both the initial predictor model and subsequent additional samples. We measure uncertainty of a sample as the cosine similarity to the class prototype. Second, to prevent sampling bias and maintain diversity, we use the class prototypes as a proxy for k-means/cluster based methods, thereby eliminating the need for a costly clustering step. We demonstrate that, remarkably, by ensuring hyperspherical uniformity of samples and using the same as a basis for diverse selection, we outperform over various active learning methods across five benchmark datasets. Our work presents the following key contributions: 

\begin{itemize}
    \item Incorporate fixed equiangular hyperspherical points as class prototypes during the active learning training process.
    \item Propose maximally separated active learning (MSAL), a novel combined uncertainty and diversity based-method for active learning on the hypersphere. Based on the cosine similarity to class prototypes, we introduce an uncertainty sampling strategy, MSAL and a diversity-based combined strategy, MSAL-D. 
    \item Showcasing state-of-the-art performance across multiple benchmark datasets in standard AL setting. Showing the ease of integrating existing AL method into our proposed method for boosted performance.
\end{itemize}

The rest of the paper is organized as follows. In Section~\ref{sec:related_work}, we present an overview of the relevant literature in both active learning and hyperspherical learning. In Section~\ref{sec:method}, we briefly discuss maximum separation for supervised learning and introduce our proposed method which leverages maximum separation for active learning. In Section~\ref{sec:experiments}, we present experiment results along with empirical analysis. We finally provide a conclusion for our study in Section~\ref{sec:conclusion}.

\section{Related Work}
\label{sec:related_work}

\subsection{Active Learning}

Active learning has been widely utilized in various fields, including image recognition \cite{gal2017deep,gudovskiy2020deep,bengar2022class}, text classification \cite{schroder2020survey,zhang2017active}, visual question answering (VQA) \cite{lin2017active}, and object detection \cite{feng2019deep, haussmann2020scalable}, and so on. 

Active learning methods are categorized based on their query strategy by which unlabled data is queried and selected for annotation \cite{tharwat2023survey}. \textbf{Information-based methods} search for most informative points by looking for most uncertain points that are expected to lie closer to the decision boundaries. These include uncertainty sampling \cite{sharma2017evidence,balcan2007margin,lewis1995sequential, seung1992query}, expected model change \cite{zhang2017active, vezhnevets2012active, freytag2014selecting,Ash2020Deep} and expected error reduction \cite{roy2001toward, yoo2019learning, zhao2021efficient}. \textbf{Representation-based methods} use the structure of unlabeled data to find the most representative points of the input space. Core-set \cite{sener2018active, mahmoodlow, yehuda2022active}, cluster-based \cite{ienco2013clustering,citovsky2021batch,yang2021batch}, diversity-based \cite{brinker2003incorporating} and density-based \cite{wu2006sampling,fu2013survey} approaches find representative samples of the unlabeled data. Several studies \cite{zhdanov2019diverse,kim2020deep,song2019combining,yang2017suggestive} have also combined the two strategies, named \textbf{combined strategy}, to select both highly uncertain and highly representative labeled samples. Other AL methods include bayesian strategies \cite{zhao2021uncertaintyaware,woo2023active, kirsch2019batchbald}, adversarial strategies using adversarial attacks \cite{ducoffe2018adversarial} or generative adversarial networks (GANs) \cite{zhu2017generative, sinha2019variational, shui2020deep}. For a thorough classification of AL methods, we refer the reader to the survey by Zhan et.al. \cite{zhan2021comparative}.

Our proposed method also falls under the combined strategy. We propose an uncertainty metric based on distance of the unlabeled samples to class prototypes on hypersphere to obtain the uncertain samples. We then select the representative samples using the similarity of these uncertain samples with the class prototypes thereby ensuring diversity of the selected samples. It is worth nothing that our proposed method is agnostic to the choice of uncertainty sampling strategy and we show that it can be replaced with any other strategies.

\subsection{Hyperspherical Learning}

Hyperspherical learning has found various applications in deep learning ranging from classification \cite{liu2018learning, lin2020regularizing, zhou2022learning, pmlr-v130-liu21d, saadabadi2024hyperspherical}, open-set learning \cite{ming2023cider,PALM2024,cevikalp2024deep, bai2024hypo} to self-supervised learning\cite{wang2020understanding,hua2021feature, jha2024common, zheng2022self,cosentino2022toward,durrant2022hyperspherically}. By means of enforcing angular separation through cosine similarity which encourages feature disentanglement, makes hyperspherical learning an ubiqutous tool in various deep learning paradigms. 

\textbf{Hyperspherical uniformity.} Uniformly distributing points on a hypersphere has been extensively studied in in physics \cite{thomson1904xxiv} and mathematics \cite{cohn2007universally} which also inspired its applications in machine learning, especially related to classification. Uniformly distributing classes on an n-dimensional hypersphere allows to learn well discriminated embeddings which was explored extensively through additional losses and objectives in softmax cross entropy optimization ~\cite{liu2018learning,lin2020regularizing,liu2016large, liu2017deep,zhou2022learning}.  On the contrary, without additional lossses, hyperspherical uniformity can be approximated by minimizing the maximum cosine similarity between pairs of class vectors to be used as class prototypes when learning representations \cite{mettes2019hyperspherical, Wang2020MMARD}. A closed-form solution also exists for hyperspherical uniformity, a simplex, and is leveraged by many recent works \cite{mroueh2012multiclass,pernici2021regular,zhu2021geometric, Kasarla2022,yang2022inducing} to replace trainable classification layer by fixed class prototypes. 

\textbf{Hyperspherical active learning.} Henrich and Obermayer \cite{henrich2008active,adiloglu2008geometrical} give performance estimates for binary active learning using a simplex constructive approach on the hypersphere. Recent work by Cao et.al., MHEAL\cite{cao2023almhe} is the equivanlent of k-means which optimizes hyperspherical energy\cite{liu2018learning} to obtain spherical k-means for representative sampling. In this work, we suggest that hyperspherical uniformity can provide a solid foundation for learning effective representations in active learning.

\section{Method}
\label{sec:method}

In this section, we first provide an overview of supervised learning with maximum separation \cite{Kasarla2022} in Section~\ref{subsec:maxsep}. We then detail our proposed hybrid active learning approach in Section~\ref{subsec:approach}. It incorporates the class prototypes obtained from the aforementioned maximum separation to learn well-discriminated class embeddings throughout the active learning training process.

\subsection{Supervised Learning with Maximum Separation}
\label{subsec:maxsep}
While the general problem of optimally separating classes on arbitrary numbers of dimensions on a hypersphere is an open problem~\cite{liu2018learning,mettes2019hyperspherical}, Kasarla et.al. \cite{Kasarla2022} propose a constructive approach that maximally separates $C$ classes on $C-1$ dimensional hypersphere. To achieve this, let $\hat{\mathbf{x}}_i = \Phi(\mathbf{x}_i) \in \mathbb{R}^{C-1}$ represent the output of the network in a $(C-1)$-dimensional output space. The class logits in the $C$-dimensional space are then obtained through the matrix multiplication: $\mathbf{o}_i = P^T \hat{\mathbf{x}}_i$, where $P \in \mathbb{R}^{(C-1) \times C}$. The matrix $P$ denotes a set of $C$ vectors, one for each class, each of $(C-1)$ dimensions. This matrix is fixed such that the class vectors are separated both uniformly and maximally, which in turn is embedded on top of a network. We outline the implementation of class vectors on a hypersphere with maximum separation below. The number of classes is denoted by $k+1 = C$ for notational convenience. All class vectors are equally separated with at an angle of $-1/k$.

\begin{align}
\nonumber 
P_1&=\begin{pmatrix}1&-1\end{pmatrix} \; \; \in \; \mathbb R^{1\times2}\\ 
P_k&=\begin{pmatrix}1&-\frac{1}{k}\mathbf1^T\\ \mathbf0&\sqrt{1-\frac1{k^2}}\,P_{k-1}\end{pmatrix} \; \in \; \mathbb R^{k\times(k+1)}
\label{eq:main}
\end{align}

Remarkably, incorporating this fixed matrix as the last layer of the neural network leads to significant performance improvements across various classification tasks. The straightforward implementation allows for seamless integration into existing active learning methods.

\textbf{Model.} For a deep network $\Phi$, we embed the fixed class prototypes, $P_c$, obtained from  maximum separation after the final layer of the network. The final class logits are obtained as $\mathbf{o}_i = \rho P_c^T \Phi(\mathbf{x}_i)$. For simplicity, we will refer to the model with maximum separation as $\Phi' = \rho P_c^T \Phi$, where $\rho$ is the radius of hypersphere. We consider $\rho = 1$ unless otherwise specified. The model is trained with a softmax cross-entropy loss. The class prototype matrix $P^T_c$ remains fixed throughout training, ensuring that the distances between class prototypes remains maximized. For each sample $\mathbf{x}_i$, the logits of the networks $\mathbf{o}_i = \Phi'(\mathbf{x}_i)$ aligns with the prototype of its corresponding class, ensuring optimal class separation.

\subsection{Maximally Separated Active Learning}
\label{subsec:approach}

This section introduces the maximally separated active learning (MSAL) method for batch-based active learning. In pool-based active learning, we are given a set of unlabeled samples denoted by $\mathcal{U}$. The goal is to select a batch of $b$ informative samples for querying from from the unlabeled set $\mathcal{U}$ with a query budget of $q$, where the final query set is $\mathcal{Q} \subset \mathcal{U}$. In each active learning round, the model, $\Phi' = P^T_c\Phi$ is trained with maximum separation. At the end of each round, the unlabeled samples are queried for labeling with the strategy outlined below. 

\textbf{Uncertainty sampling.} To find most informative points closer to the decision boundaries, we use the model's inherent clusters around the class prototypes as the basis for determining uncertainty. Given the trained representations from the model on the starting set, $\Phi'$, for each of unlabeled images $\mathbf{x}_i \in \mathcal{U}$, the model features are $\mathbf{o}_i = \Phi'(\mathbf{x}_i)$. The closest prototype to the feature is established and this distance denotes a quantification of uncertainty: if sufficiently close, the test sample is highly likely to belong the the respective class; if too distant, then the test sample is uncertain. We rank all the unlabeled samples in ascending order according to the distance to the closest class prototype, obtained by the cosine similarity of the model features to the prototype, defined as,

\begin{align}
\label{eq:ms-uncertain}
    \alpha_{\text{MSAL}}^i = \max_c \frac{\hat{\mathbf{x}}_i \cdot P^T_c}{\max\norm{\hat{\mathbf{x}}_i}_2}
\end{align}

Since the model is trained to maximize the alignment between samples and their fixed class prototypes, using Equation~\ref{eq:ms-uncertain}, we can determine the samples that are farthest from the class prototypes and are thus the most uncertain.

\textbf{Diversity sampling.} We pre-filter top $\beta b$ uncertain examples where $\beta$ is a hyperparameter. This step is similar to that of Zhdanov et.al., \cite{zhdanov2019diverse}. We consider the class prototypes as cluster centers. We then sample $b/C$ closest samples to each cluster center from the $\beta b$ samples. This ensures to give both representative and diverse samples from the unlabeled pool.

The algorithm is summarized in Algorithm~\ref{alg:main-algo}. 

\begin{algorithm}[H]
    \algsetup{linenosize=\tiny}
    \small
    \caption{Maximally separated active learning with diversity sampling (MSAL-D)}
    \label{alg:main-algo}
    \begin{algorithmic}
        \STATE {\bfseries Input:} \\
        $\mathcal{L}_0, \mathcal{U}_0$ : Initial labeled and unlabeled samples\\
        $b, q$ : batch and query size\\
        $T$ : number of acquisition steps\\        
        \FOR {$t=0$ {\bfseries to} $T-1$}
        \STATE Obtain $\Phi' = P^T_c \Phi$ by training on $\mathcal{L}_t$
        \STATE Compute uncertainty of unlabeled samples with   $\alpha_{\text{MSAL}}^i = \max_c \frac{\hat{\mathbf{x}}_i \cdot P^T_c} {\max\norm{\hat{\mathbf{x}}_i}_2}$ 
        \STATE Sample $\beta b$ most uncertain samples
        \FOR {$c=0$ {\bfseries to} $C-1$}
        \STATE find $b/C$ closest samples, $\mathbf{x}_c = {\argmax}_c \alpha_{\text{MSAL}}$ for each class $C$
        \STATE Update the query set, $\mathcal{Q}_c \gets \mathcal{Q}_{c-1} \cup \{\mathbf{x}_c\}$
        \ENDFOR
        \STATE update the labeled samples, $\mathcal{L}_{t+1} \gets \mathcal{L}_t \cup \{(\mathbf{x}_i, \mathbf{y}_i)\}_{\mathbf{x}_i \in \mathcal{Q}_q}, \ \mathcal{U}_{t+1} \gets \mathcal{U}_t \setminus \mathcal{Q}_q$
        \ENDFOR
    \end{algorithmic}
\end{algorithm}

\section{Experiments}
\label{sec:experiments}

This section presents the empirical evalution of the proposed active learning approach. Section~\ref{sec:setup} gives an detailed overview of datasets and hyperparameters. In Section~\ref{sec:results}, we present our results, empirical analysis and hyperparameter ablations.

\subsection{Setup}
\label{sec:setup}

\textbf{Datasets.} We compare our methods performance against various uncertainty-based active learning algorithms on diverse benchmark datasets: MNIST\cite{deng2012mnist}, SVHN \cite{netzer2011reading}, CIFAR\cite{krizhevsky2009learning} and TinyImagenet\cite{le2015tiny}. For the imbalanced learning experiments, we use the long-tailed classification setting of CIFAR-10 and 100 datasets \cite{cui2019class} with imbalance factors  $0.1$ and $0.01$. The imbalance factor represents the ratio between the number of samples in the least frequent class and the number of samples in the most frequent class within the dataset.

\textbf{Methods.}  We compare our proposed method with uncertainty-based, diversity-based and combined methods. Uncertainty-based methods include Least Confidence\cite{wang2014new}, Margin Sampling,\cite{netzer2011reading} Entropy Sampling \cite{shannon2001mathematical} and their MC-dropout variants\cite{beluch2018power}, BALD\cite{gal2017deep}, Mean Standard Deviation\cite{kampffmeyer2016semantic}, Variation Ratios \cite{freeman1965elementary} and CEAL(Entropy variant)\cite{wang2016cost}. Representatative/Diversity based methods include the K-Means, K-Means(GPU) implemented by \DeepALPlus\cite{zhan2022comparative} and KCenter\footnote{Note that we don't report Coreset\cite{sener2018active}, similar to \DeepALPlus because Coreset uses the Gourbi optimizer to solve for the KCenter problem, which is not a free optimizer. Instead we use KCenter from \DeepALPlus as a greedy optimization of Coreset approach.}. Since our approach is a combined stragegy, we also compare to combined methods like BADGE\cite{ash2019deep} and DBAL\cite{zhdanov2019diverse} which combine strategies from uncertainty and representative-based methods. We also exclude comparisons to any adversarial-based methods \cite{shui2020deep, sinha2019variational, ducoffe2018adversarial,zhu2017generative} to maintain fair comparison since our method doesn't have any adversarial learning. 

\textbf{Implementation details.} For all the reported AL methods, we employed ResNet18 (w/o pre-training) \cite{he2016deep} as the basic learner. For a fair comparison, consistent experimental settings of the basic classifier are used across all AL methods. We run these experiments using \DeepALPlus toolkit\cite{zhan2022comparative}. The settings are as follows: for MNIST and SVHN, ResNet18 has kernel size of $7\times7$ in the first convolutional layer and rest have $3\times3$ kernel size. We adopt Adam optimizer \cite{kingma2014adam} with learning rate $0.001$ for all the experiments. Number of training epochs in each active learning round for MNIST and SVHN is 20, for CIFAR-10 and CIFAR-100 is 30 and for TinyImageNet is 40 respectively. We summarize the query set and batch size in Table~\ref{tab:impl}. We conduct all the experiments on a single NVIDIA A100-SXM4 GPU with 40GB memory.

\begin{table}[ht]
    \centering
    \caption{\textbf{Dataset setting in the AL methods.} \(\#i\) is the size of initial labeled pool, \(\#u\)is the size of unlabeled data pool \(\#e\) is number of epochs used to train the basic classifier in each AL round. \(b\) is the batch size of queried labeled samples in each AL round and \(Q\) is the total query budget.
}
    \resizebox{7cm}{!}{
    \begin{tabular}{l c c c c c }
    \toprule
    Dataset& \(\#i\) &  \(\#u\)&\(b\) &\(Q\) & \(\#e\) \\
    \midrule
        MNIST & 500 & 59,500 & 250 & 10,000 & 20  \\
        SVHN & 500 &72,757  & 250 & 20,000 & 20 \\
        CIFAR-10 & 1,000 & 49,000 & 500 & 40,000 & 30 \\
        CIFAR-100 & 1,000 & 49,000 & 500 & 40,000 & 40 \\
        TinyImageNet & 1,000 & 99,000 & 500 & 40,000 & 40 \\
        \bottomrule
    \end{tabular}
    }
    \label{tab:impl}
\end{table}

\textbf{Evaluation metrics.} We run the experiments with random splits of the initial labeled and unlabeled sets for three runs and report the average performance. We report the \textit{overall performance} using area under the budget curve (AUBC) \cite{zhan2021multiple,zhan2021comparative}. AUBC is computed by evaluating the AL method for different fixed budgets (e.g., accuracy vs budget) and higher AUBC indicates better overall performance.  We also report the final accuracy (F-acc) after the budget $\mathcal{Q}$ is exhausted.

\subsection{Results}
\label{sec:results}

\textbf{Comparison to active learning approaches.} We compare our method to various active learning approaches and report the results in Table~\ref{tab:main-results}. We divide the comparisons into three different types: \textit{Cmb vs Ours}, \textit{Div vs Ours} and \textit{Unc vs Ours}. \textit{Cmb vs Ours:} Our proposed method (MSAL-D) is a combined strategy, without the expensive clustering step of BADGE and DBAL. We outperform DBAL across all datasets, and BADGE on MNIST, SVHN and TinyImagenet datasets. Most notable is TinyImagenet, where the clustering based combined strategies struggle to get good performance, but our method outperforms by a large margin. \textit{Unc vs Ours:} Our uncertainty metric, MSAL, is on-par with other uncertainty based metrics. Our proposed combined strategy outperforms all uncertainty based strategies on MNIST, and TinyImageNet datasets, and in the AUBC metric on SVHN dataset. 
\textit{Div vs Ours:} Diversity/representative-based clustering strategies like KMeans, KMeans(GPU) and KCenter perform poorer in comparison to both uncertainty based and ours. 

\begin{table*}[ht]

\centering
\caption{\textbf{Overall results.} We compare our proposed method to different active learning strategies from uncertainty-based, representative-based, and combined strategies. In our methods, MSAL is uncertainty only metric computed from Equation~\ref{eq:ms-uncertain} and MSAL-D also includes diversity based sampling as summarized in Algorithm~\ref{alg:main-algo}. \textbf{bold} represents best performance and \underline{underlined} represents second best performance.  $\dag$ reported from survey paper by Zhan et.al.\cite{zhan2022comparative}}. \label{tab:main-results}
\resizebox{12cm}{!}{
\begin{tabular}{ll|cc|cc|cc|cc|cc}
\hline 
&& \multicolumn{2}{c|}{\emph{MNIST}}  & \multicolumn{2}{c|}{\emph{SVHN}} & \multicolumn{2}{c|}{\emph{CIFAR10}} & \multicolumn{2}{c|}{\emph{CIFAR100}}  & \multicolumn{2}{c}{\emph{TinyImageNet}} \\

&Model & AUBC & F-acc & AUBC & F-acc & AUBC & F-acc & AUBC & F-acc & AUBC & F-acc \\
\hline 
&Full & $-$ & $0.9916$ & $-$ & $0.9190$ & $-$ & $0.8793$ & $-$ & $0.6062$  & $-$ & $0.4583$ \\
&\textbf{Random} & $ 0.959$ & $0.9814$ & $0.8337$ & $0.8995$ & $0.7970$ & $0.8670$ & $0.468$ & $0.5895$ & $0.2573$ & $0.3617$ \\

\Xhline{0.6pt}

\multirow{10}{*}{\rotatebox{90}{Unc}} &\textbf{LeastConf} & $0.9710$ & $0.9868$ & $0.8653$ & $0.9035$ & $0.8137$ & $0.8729$ & $0.4750$ & $0.5982$ & $0.2447$ & $0.3388$ \\

&\textbf{LeastConfD}$^\dag$ & $0.9750$ & $0.9915$ & $0.8320$ & $0.9083$ & $0.8137$ & $0.8825$ & $0.4730$ & $0.5997$ &  $0.2620$ & $0.3698$ \\

&\textbf{Margin} & $0.9723$ & $0.9903$ & $0.8637$ & $0.9286$ & $\underline{0.8147}$&$0.8815$&$0.4777$&$0.5998$&$0.2553$ & $0.3573$  \\

&\textbf{MarginD}$^\dag$ & $0.9703$& $0.9899$ & $0.8357$ & $0.9104$ &  $0.8140$ & $\mathbf{0.8837}$ & $0.4777$ & $0.6000$ &  $0.2607 $ & $0.3541$  \\

&\textbf{Entropy} & $0.968$ & $0.9917$  & $0.862$ & $\mathbf{0.9316}$ & $0.8120$ & $\underline{0.8828}$ & $0.4680$ & $\mathbf{0.6049}$ &  $0.2357$ & $0.3332$   \\

&\textbf{EntropyD}$^\dag$ & $0.9683$ & $0.9887$ & $0.8290$ & $0.9091$ & $0.8140$ &  $0.8787$  & $0.4677$ & $0.6004$ &  $0.2627 $ & $0.3716$  \\

&\textbf{BALD}$^\dag$ & $0.9697$ & $0.9885$ & $0. 8333$ &$0.9020$ &$0.8103$ &$0.8762$&$0.4760$ &$0.5942$& $0.2623 $ &$0.3648$  \\

&\textbf{MeanSTD}$^\dag$ & $0.9713$ & $0.9735$ & $0.8323$ & $0.9087$ & $0.8087$ & $0.8821$ & $0.4717$ & $0.5963$ &  $0.2510$ & $0.3551$    \\

&\textbf{VarRatio}$^\dag$ & $0.9717$ & $0.9841$ & $0.8357$ & $0.9079$ & $\mathbf{0.8150}$ & $0.8780$ & $0.4747$ & $0.5959$ &  $0.2407 $ & $0.3426$   \\

&\textbf{CEAL(Entropy)} & $0.9787$ & $0.9889$ & $0.8430$ & $0.9142$ & $0.8143 $ & $0.8769$ & $0.4693$ & $\underline{0.6043}$ &  $0.2347$ & $0.3400$   \\

\Xhline{0.1pt}

\multirow{3}{*}{\rotatebox{90}{Div}} &\textbf{KMeans}$^\dag$ & $0.9640$ & $0.9813$ & $0.8027$ & $0.8671$ & $0.7910$ & $0.8713$ & $0.4570$ & $0.5834$ &   $0.2447 $ & $0.3385$   \\

&\textbf{KMeans (GPU)}$^\dag$ & $0.9637$ & $0.9747$ &  $0.8120$ & $0.8688$ &  $0.7977$ & $0.8718$ &  $0.4687$ & $0.5842$  &  $0.1340$ & $0.2288$  \\

&\textbf{KCenter}$^\dag$ & $0.9740$ & $0.9877$ & $0.8741$ & $0.8283$ & $0.8047$ & $0.8741$ & $0.4770$&$0.5993$ &$0.2540$ & $ 0.346$  \\

\Xhline{0.1pt}

\multirow{2}{*}{\rotatebox{90}{Cmb}}&\textbf{BADGE(KMeans++)}$^\dag$ & $0.9707$ & $0.9904$ & $0.8377$ & $0.9057$ & $0.8143$ & $0.8794$  & $\mathbf{0.4803}$ & $0.6034$ &  - &- \\

&\textbf{DBAL(KMeans)} & $0.962$ & $ 0.985$ & $0.8317$ & $0.8941$ & $0.7977$ & $0.8736$ & $0.4687$& $0.5938$ &  $0.159$& $0.2488$ \\

\Xhline{0.1pt}

\multirow{2}{*}{\rotatebox{90}{Ours}} &\textbf{MSAL} & $0.9760$ & $0.9924$  & $0.8653$ & $\underline{0.9306}$  & $0.8127$&$0.8792$ &$0.4710$ &$0.6008$ &  $0.2443$ & $0.3434$  \\

&\textbf{MSAL-D} &$\textbf{0.9820}$ & $\textbf{0.9931}$ & $\textbf{0.8657}$ & $0.9287$ & $0.8107$ & $0.8754$ &   \underline{0.4780} &  0.5975 &  $\mathbf{0.2717}$ & $\mathbf{0.3741}$  \\
\hline 
\end{tabular}
}

\label{performance}
\end{table*}

\begin{figure}[ht]
    \centering
      \includegraphics[width=0.35\linewidth]{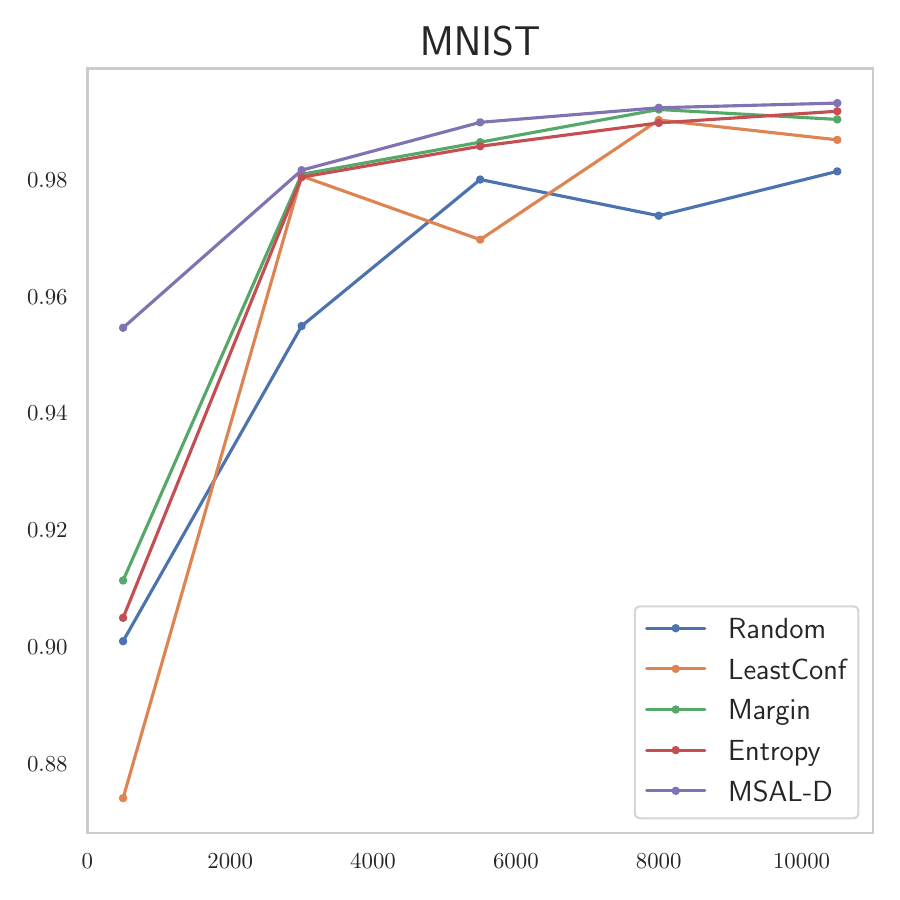}
    \includegraphics[width=0.35\linewidth]{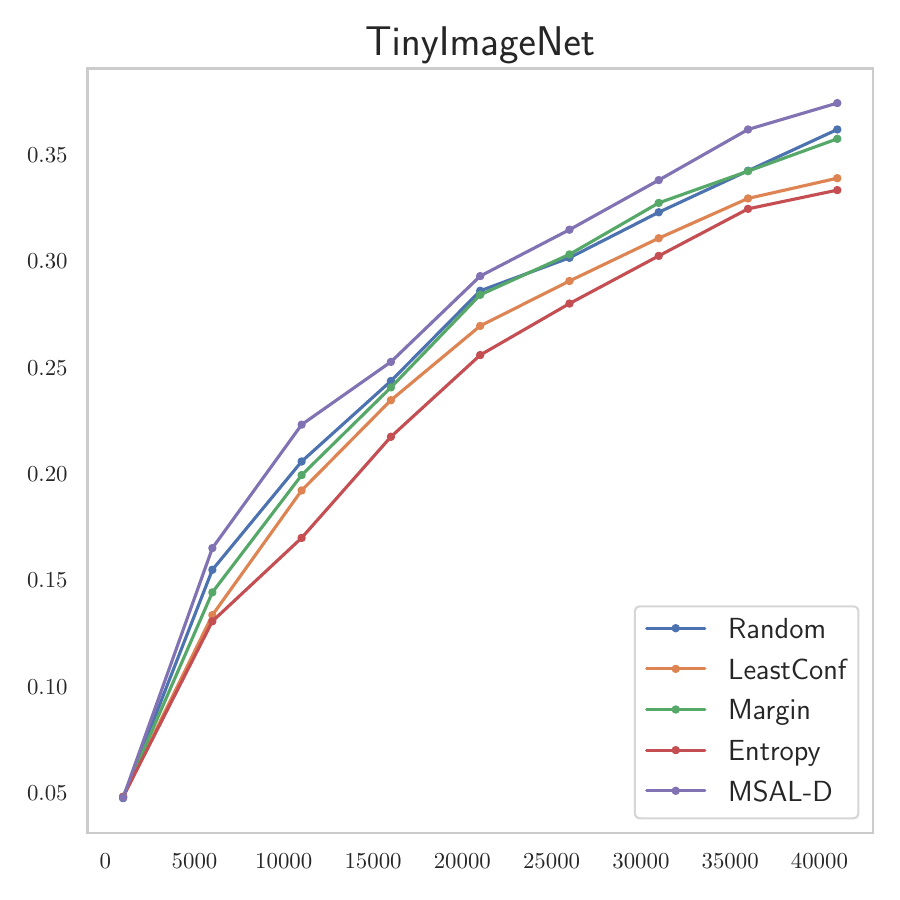} 

    \caption{\textbf{Overall accuracy vs. budget curves} on \emph{MNIST} and \emph{TinyImageNet} datasets.}
    \label{fig:enter-label}
\end{figure}

\textbf{Maximum separation can also be added to existing uncertainty based approaches.} In this setting, the network is $\Phi'$ with maximum separation. We replace our uncertainty strategy with other strategies like Least Confidence, Entropy and Margin for experiments with (MSAL), and add the diversity sampling for (MSAL-D). Across all settings, as reported in Table~\ref{tab:uncertain-msal}, we notice that making the active learning approach maximally separated boosts the performance across the datasets in both AUBC and F-acc metrics.

\begin{table}[ht]
\centering
\caption{\textbf{Uncertainty approaches with maximum separation.} We replace our uncertainty strategy with other strategies like Least Confidence, Entropy and Margin for experiments with (MSAL), and add the diversity sampling for (MSAL-D). We show that our proposed approach with maximum separation is versatile to be used in any other uncertainty metric with good performance gains. \textbf{bold} represents best performance and \underline{underlined} represents second best performance. }\label{tab:uncertain-msal}
\resizebox{\linewidth}{!}{
\begin{tabular}{ll|cc|cc|cc|cc|cc}
\hline 
\multicolumn{2}{c|}{}& \multicolumn{2}{c|}{\emph{MNIST}}  & \multicolumn{2}{c|}{\emph{SVHN}} & \multicolumn{2}{c|}{\emph{CIFAR10}} & \multicolumn{2}{c|}{\emph{CIFAR100}}  & \multicolumn{2}{c}{\emph{TinyImageNet}} \\
Model & & AUBC & F-acc & AUBC & F-acc & AUBC & F-acc & AUBC & F-acc & AUBC & F-acc \\
\hline
Full && $-$ & $0.9916$ & $-$ & $0.9190$ & $-$ & $0.8793$ & $-$ & $0.6062$  & $-$ & $0.4583$ \\
\textbf{Random} && $ 0.959$ & $0.9814$ & $0.8337$ & $0.8995$ & $0.7967$ & $0.8760$ & $0.468$ &$0.5895$ & $0.2573$ & $0.3617$ \\
\Xhline{0.1pt}
\textbf{LeastConf} && $0.9710$ & $0.9868$ & $\mathbf{0.8653}$ & $0.9035$ & $0.8137$ & $0.8729$ & $\underline{0.4750}$ &$\underline{0.5982}$ & $\underline{0.2447}$ & $ 0.3388$ \\
\rowcolor{Gray} 
\textbf{LeastConf(MSAL)} && $\underline{0.9770}$ & $\mathbf{0.9922}$ & $\underline{0.8623}$ & $\underline{0.9263}$ & $\mathbf{0.8143}$ & $\mathbf{0.8825}$ & $0.4743$ &$\mathbf{0.606}$ &$0.243$& $\underline{0.3449}$ \\
\rowcolor{Gray0} \textbf{LeastConf(MSAL-D)} && $\mathbf{0.9810}$ & $\underline{0.9903}$ & $0.8613$ & $\mathbf{0.9288}$ & $0.8120$ & $0.8677$& $\mathbf{0.4773}$ & $0.5915$ &$\mathbf{0.2737}$ & $\mathbf{0.3816}$ \\
\Xhline{0.1pt}
\textbf{Margin} && $0.9723$ & $0.9903$ & $\underline{0.8637}$ & $\underline{0.9286}$ & $0.8133$ & $0.8800$ & $\underline{0.4777}$ &$\underline{0.5998}$ & $\underline{0.2553}$ & $\underline{0.3573}$ \\
\rowcolor{Gray} \textbf{Margin(MSAL)} && $\mathbf{0.9820}$& $\underline{0.9913}$ & $0.8630$ & $0.9277$ & $\mathbf{0.8137}$ & $\mathbf{0.8826}$ &$\mathbf{0.4780}$ &$\mathbf{0.5999}$ & $0.2550$ & $0.3566$ \\
\rowcolor{Gray0} \textbf{Margin(MSAL-D)} && $\mathbf{0.9820}$& $\mathbf{0.9917}$ & $\mathbf{0.8647}$ & $\mathbf{0.9324}$ & $0.8107$ & $0.8722$ &$0.4770$ & $0.5996$ &$\mathbf{0.2747}$ & $\mathbf{0.3829}$ \\
\Xhline{0.1pt}
\textbf{Entropy} && $0.968$ & $\mathbf{0.9917}$  & $0.862$ & $0.9316$ & $0.8120$ & $\mathbf{0.8828}$ & $0.468$ &$\mathbf{0.6049}$ & $0.2357$ & $0.3332$ \\
\rowcolor{Gray} \textbf{Entropy(MSAL)} && $\mathbf{0.976}$ & $\mathbf{0.9917}$ & $\mathbf{0.863}$ & $\mathbf{0.9332}$ & $\mathbf{0.8130}$  &$0.876$&$\underline{0.469}$ & $0.5969$ & $\underline{0.2373}$ & $\underline{0.343}$ \\
\rowcolor{Gray0} \textbf{Entropy(MSAL-D)} && $\mathbf{0.9807}$ & $\underline{0.9909}$ & $\underline{0.8627}$ & $\underline{0.9273}$ & $0.8103$ & $ 0.8752$&$\mathbf{0.477}$ & $ \underline{0.5978}$ & $\mathbf{0.2693}$ & $\mathbf{0.3724}$ \\
\hline
\end{tabular}
}

\end{table}

\textbf{High-confidence pseudo labeling.} CEAL \cite{wang2016cost} psuedo-labels the most confident predictions, defined by high-confidence threshold, $\delta$, which reduces the total number of manual annotations. In our proposed approach, the confident predictions are the closest ones to the cluster center. Following the same approach and hyperparameter settings of CEAL for pseudo-labeling, we report the performance on MSAL with pseudo-labeling in Table~\ref{tab:ceal-ms}. With pseudo-labeling, we notice that our method outperforms CEAL with entropy sampling by a large margin across all dataset settings.  

\begin{table}[ht]
\centering
    \caption{\textbf{Pseudo-labeling confident examples} with our approach (MSAL) outperforms CEAL on all datasets.}
    \label{tab:ceal-ms}
    \resizebox{0.6\linewidth}{!}{
    \begin{tabular}{l c c c c}
    \toprule
    & \multicolumn{2}{c}{\emph{CEAL(Entropy)}} & \multicolumn{2}{c}{\emph{CEAL(MSAL)}} \\
        & AUBC & F-acc & AUBC & F-acc \\
        \midrule
       \textbf{MNIST} & $0.9787$ & $0.9889$ &$\mathbf{0.9807}$& $\mathbf{0.9928}$\\
       \textbf{SVHN} & $0.8430$ & $0.9142$& $\mathbf{0.8807}$ & $\mathbf{0.9325}$\\
       \textbf{CIFAR-10} & $0.8143$ & $0.8769$&$0.8137$&$\mathbf{0.8829}$ \\
       \textbf{CIFAR-100} &$0.4693$& $0.6043$ &$\mathbf{0.4707}$&$0.5983$ \\
       \textbf{TinyImageNet} & $0.2347$ & $0.3400$ &$\mathbf{0.2424}$&$\mathbf{0.3547}$\\
       \bottomrule
    \end{tabular}
    }
\end{table}

\textbf{Long-tailed active learning.} Learning with maximum separation is expected to improve learning especially when dealing with under-represented classes. We expect the same to hold for imbalanced active learning. By following the imbalanced data setting counterpart from classification \cite{cui2019class}, we use the imbalance factors of $0.1$ and $0.01$ for both CIFAR-10 and CIFAR-100. We report the results in Table~\ref{tab:imb-al} comparing to Least Confidence, Entropy and Margin-based sampling. MSAL outperforms on CIFAR-10 for higher imbalance of $0.01$ and is on par with other methods on imbalance of $0.1$. For CIFAR-100, MSAL is on-par with other methods on imbalance of $0.01$ but underperforms on $0.1$. While this is a promising result, a closer look into hyperparameters settings and embeddings in a future work will be beneficial to imbalanced setting of active learning.

\begin{table}[ht]
\centering
\caption{\textbf{Imbalanced Active Learning.} Reported values of imbalanced active learning for CIFAR-10 and CIFAR-100 with imbalance factors \cite{cui2019class} of $0.1$ and $0.01$.} \label{tab:imb-al}
\resizebox{\linewidth}{!}{
\begin{tabular}{lcccc}
\toprule
   & \multicolumn{2}{c}{0.1}  & \multicolumn{2}{c}{0.01} \\
CIFAR-10 & AUBC & F-acc & AUBC & F-acc \\
\midrule
Full  & $-$ & $0.8060$ & $-$ & $0.6433$  \\
\textbf{Random} & $0.6797$ &$0.7676$ & $0.5023$ &$0.5994$  \\
\midrule
\textbf{LeastConf}&  $\mathbf{0.6940}$ &$\mathbf{0.7976}$ & $0.5333$ &$0.5995$  \\

\textbf{Margin} & $0.6917$ &$0.7806$ & $\underline{0.5357}$ &$0.6126$  \\
\textbf{Entropy}&$\mathbf{0.6940}$ &$\mathbf{0.7909}$ &  $\mathbf{0.5380}$ &$0.5876$  \\
\rowcolor{Gray}\textbf{Ours(MSAL)} & $\underline{0.6930}$ &$0.7803$ & $\mathbf{0.5380}$ & $\mathbf{0.6358}$  \\

\bottomrule
\end{tabular}

\begin{tabular}{lcccc}
\toprule
  & \multicolumn{2}{c}{0.1} & \multicolumn{2}{c}{0.01} \\
CIFAR-100 & AUBC & F-acc & AUBC & F-acc \\
\midrule
Full & $-$ & $ 0.5081$ & $-$ & $0.3293$  \\
\textbf{Random} &$ 0.3233$ &$0.4344$ & $0.2267$ &$0.3125$  \\
\midrule
\textbf{LeastConf} & $0.3237$ &$0.4367$ & $0.2320$ &$0.3119$  \\

\textbf{Margin} & $\mathbf{0.3263}$ &$\mathbf{0.4413}$ & $\mathbf{0.2327}$ &$0.3080$  \\
\textbf{Entropy} & $0.3193$ &$0.4361$ & $0.2290$ &$\mathbf{0.3128}$  \\
\rowcolor{Gray}\textbf{Ours(MSAL)} & $0.3230$ &$0.4395$& $\underline{0.2320}$ &$0.3111$  \\

\bottomrule
\end{tabular}

}

\end{table}

\subsection{Analysis}

\begin{figure}[ht]\captionsetup[subfigure]{justification=centering}

    \centering
    \begin{subfigure}[b]{\textwidth}
    \centering
    \includegraphics[width=0.24\linewidth]{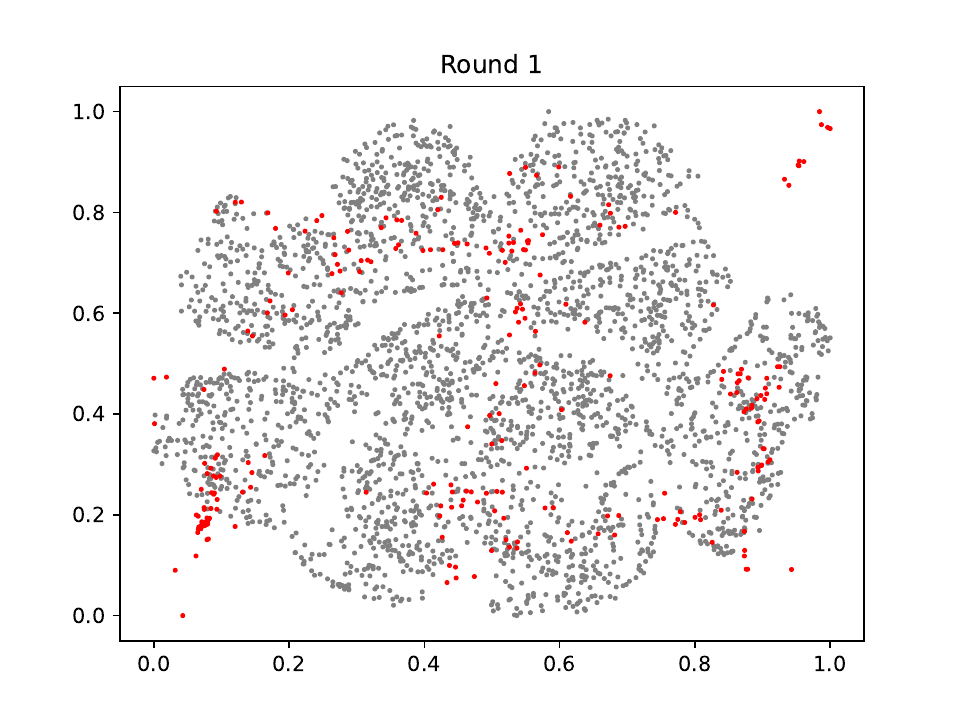}
    \includegraphics[width=0.24\linewidth]{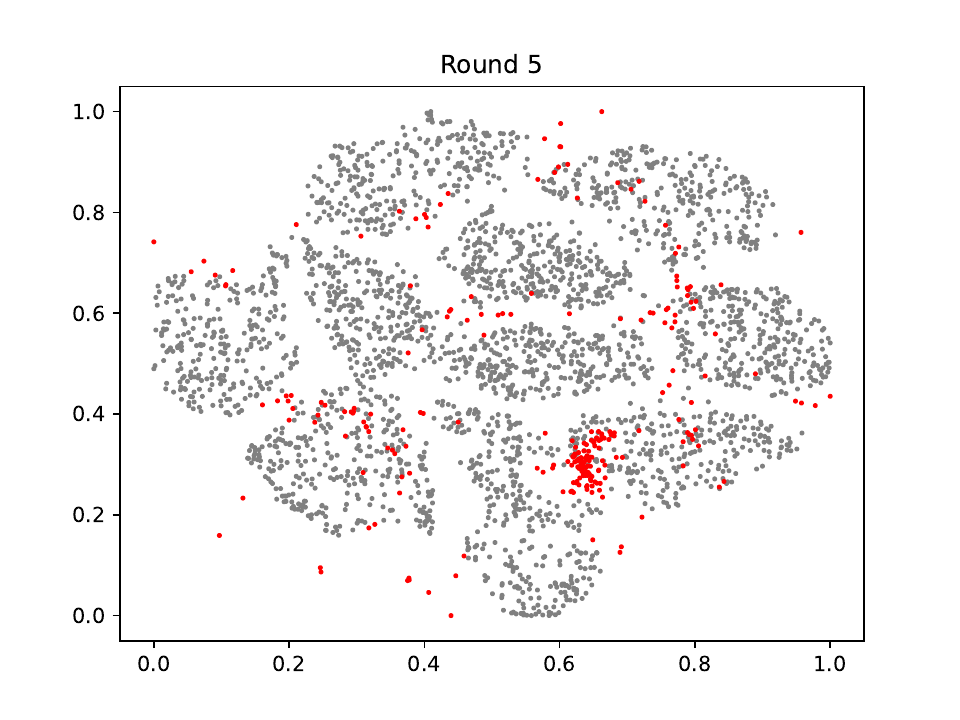}
    \includegraphics[width=0.24\linewidth]{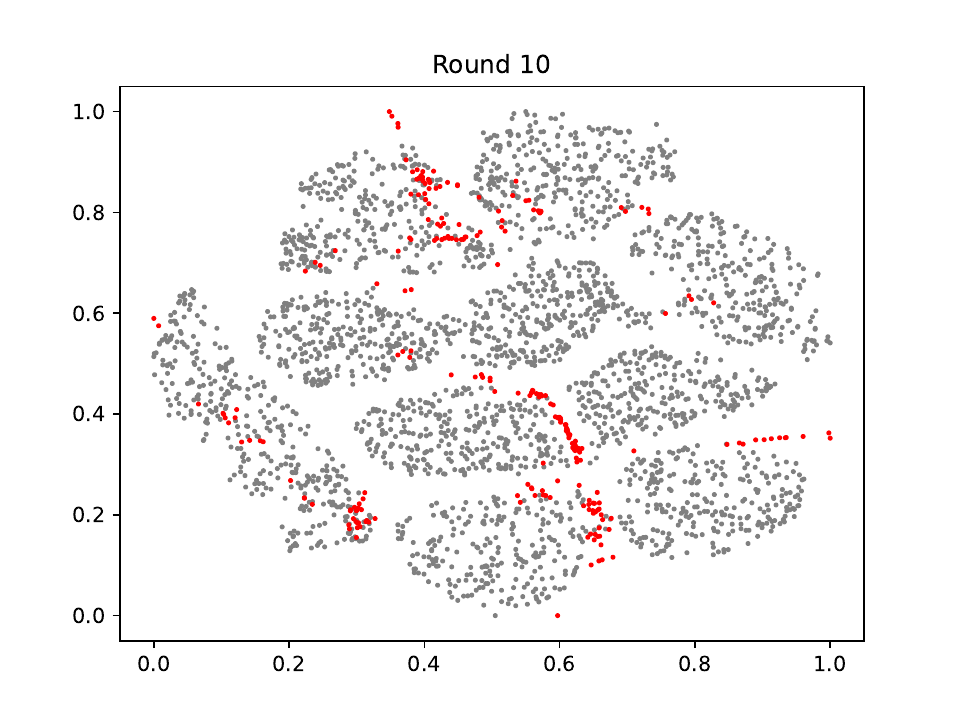}    
    \includegraphics[width=0.24\linewidth]{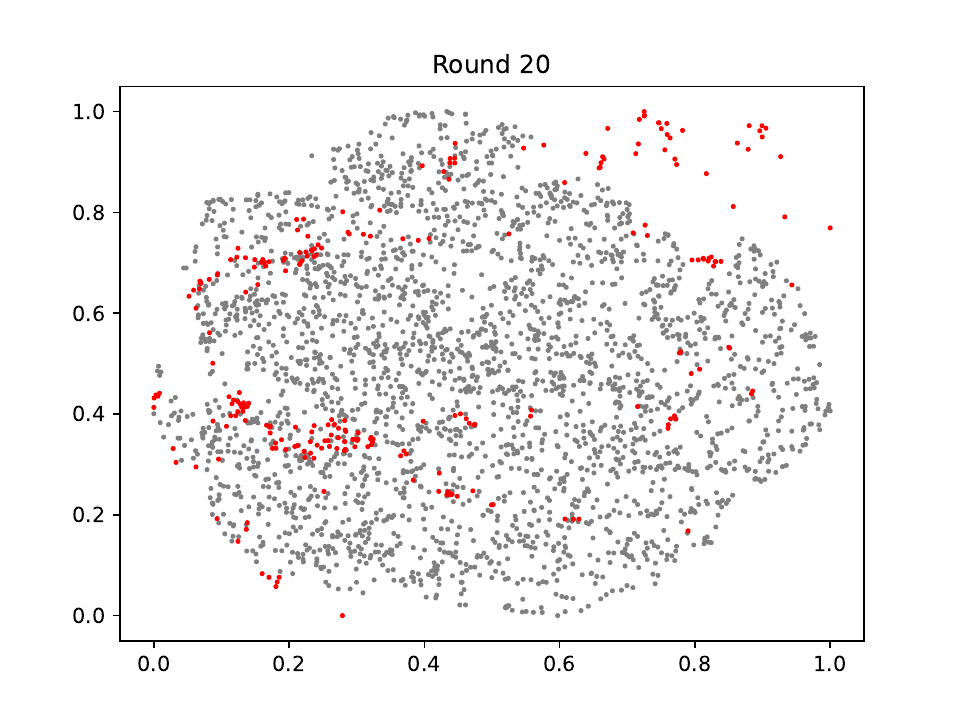}
    \caption{Entropy Sampling}\label{subfig:tsne-entropy}
    
    \end{subfigure}
    \begin{subfigure}[b]{\textwidth}
    \includegraphics[width=0.24\linewidth]{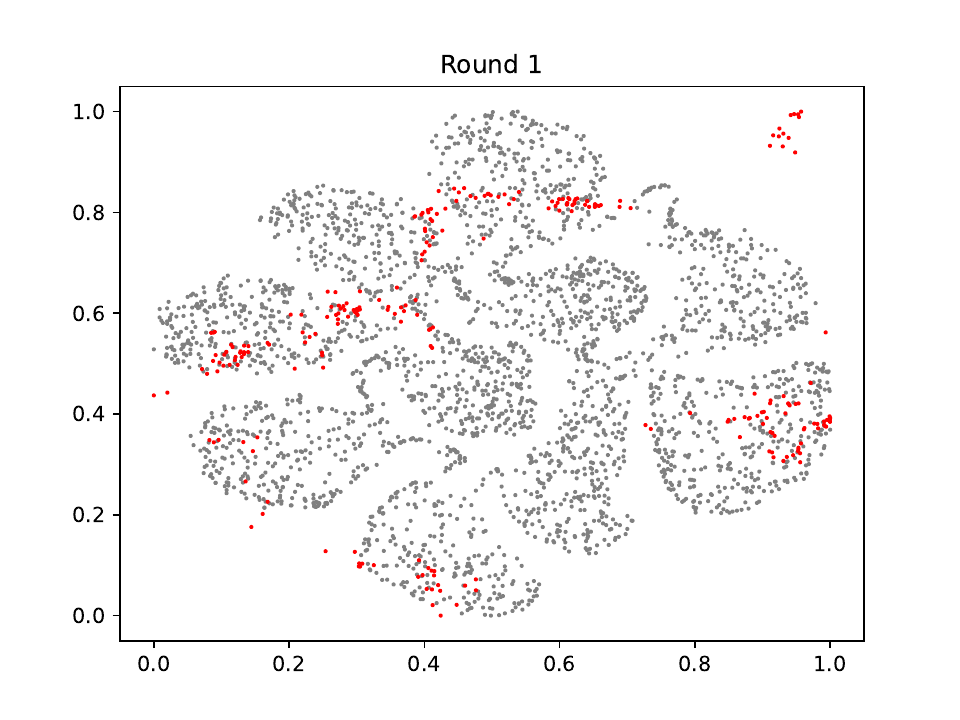}
    \includegraphics[width=0.24\linewidth]{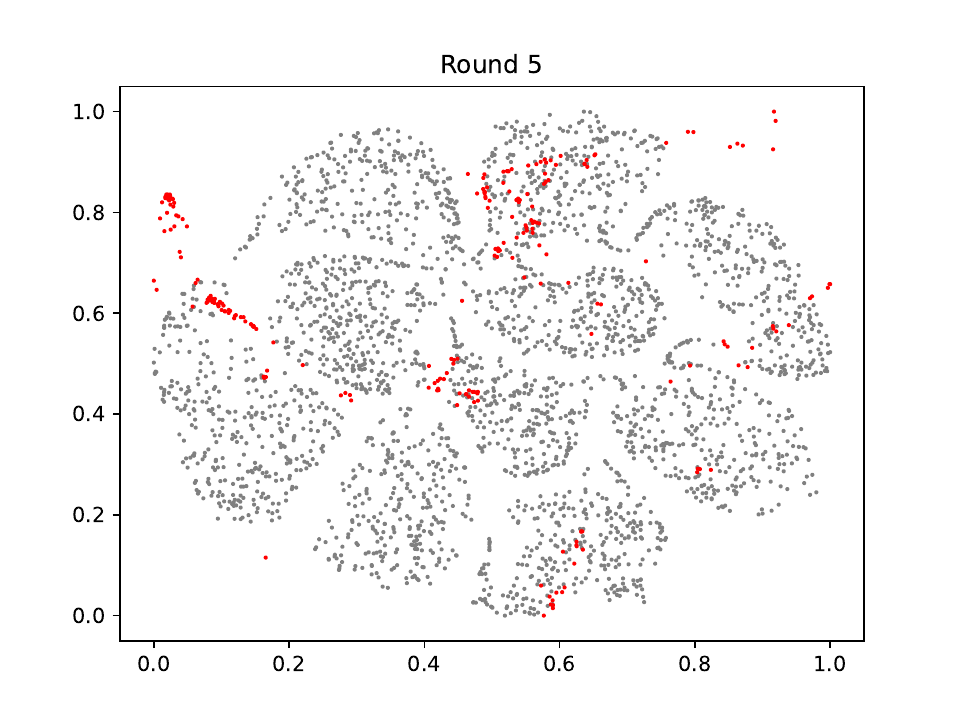}
    \includegraphics[width=0.24\linewidth]{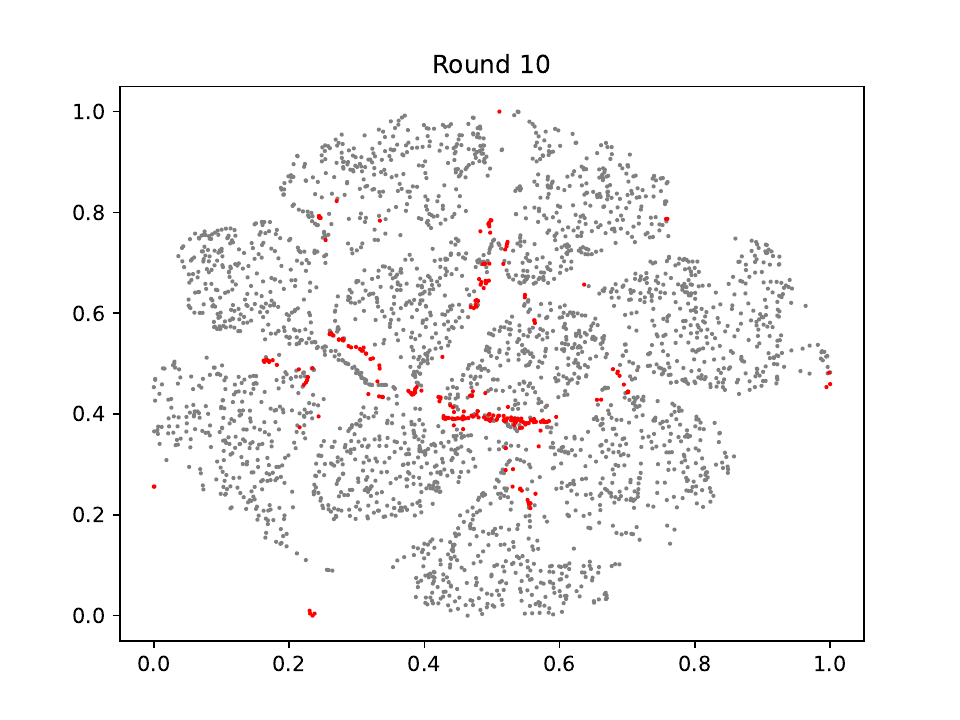}
    \includegraphics[width=0.24\linewidth]{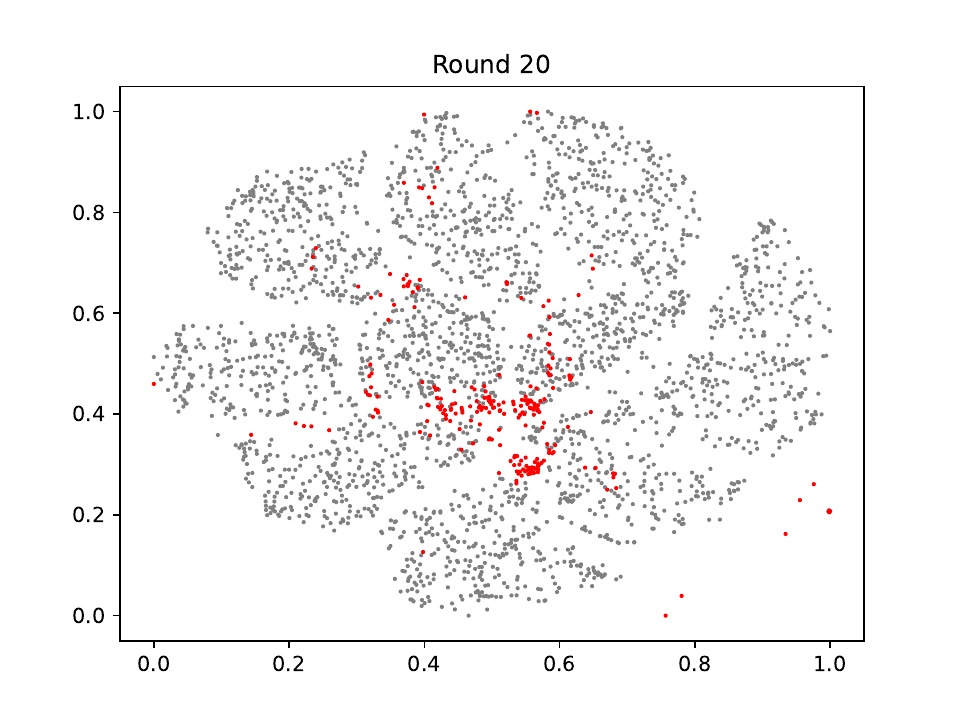}
    \caption{Ours(MSAL)}\label{subfig:tsne-msal}
    
    \end{subfigure}
    \begin{subfigure}[b]{\textwidth}

    \includegraphics[width=0.24\linewidth]{figures/tsne_msal_round1.pdf}
    \includegraphics[width=0.24\linewidth]{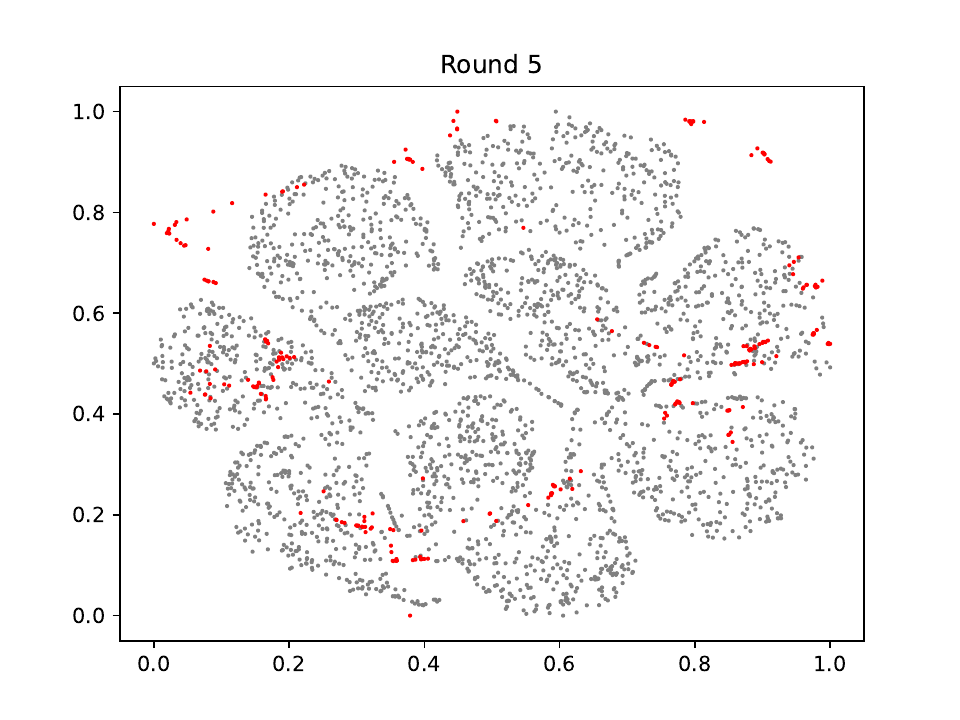}
    \includegraphics[width=0.24\linewidth]{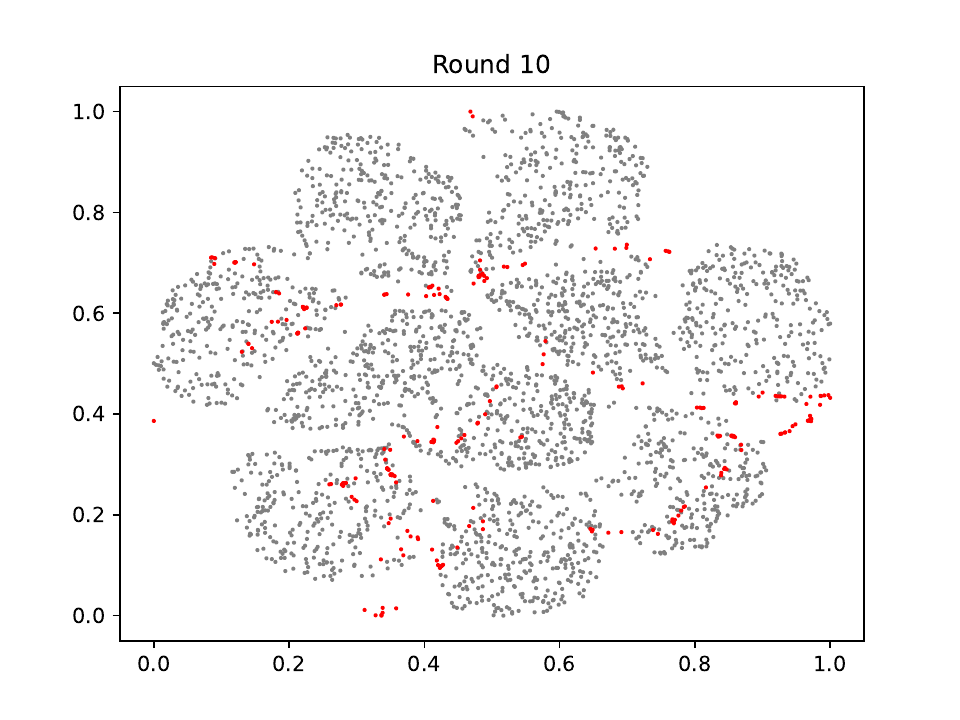}
    \includegraphics[width=0.24\linewidth]{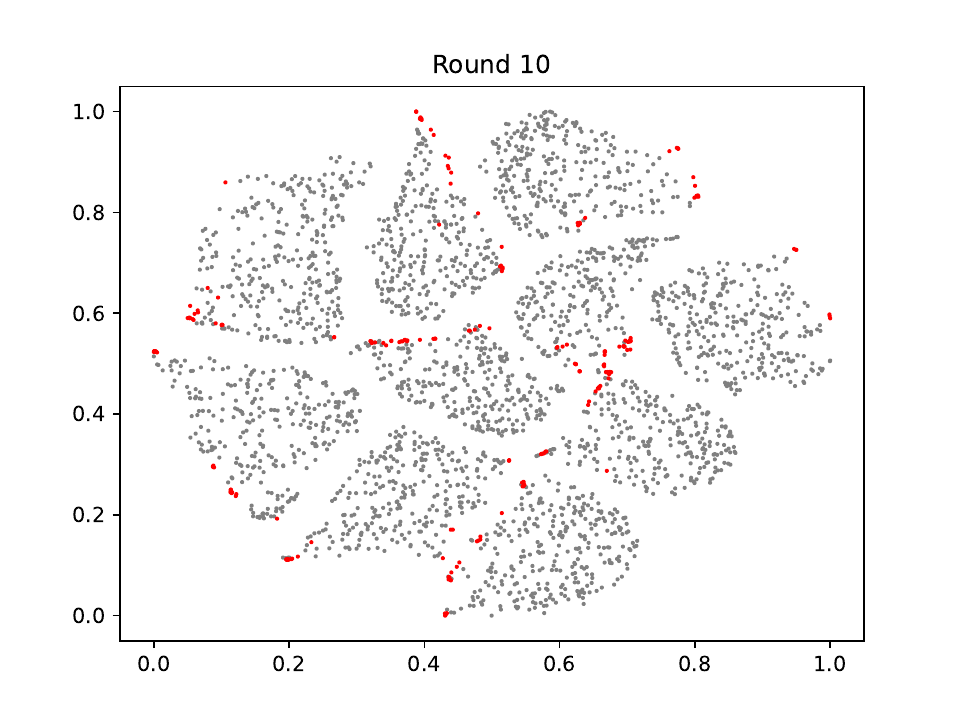}
    \caption{Ours(MSAL-D)}\label{subfig:tsne-msald}
    
    \end{subfigure}
    \caption{\textbf{A visualization of embeddings and selected samples for labeling.} We plot the t-SNE embeddings for rounds 1,5,10,20 (left $\rightarrow$ right) for entropy sampling and our methods. {\color{gray}{Grey}} points are the embeddings of the unlabeled set and {\color{red}{red}} points are the samples selected by the active learning strategy for that round. Our methods exhibit better separation with distinct clusters throughout the active learning process.}
    \label{fig:al-vis}
\end{figure}

\textbf{Learning with maximum separation creates better boundaries.} We visualize the t-SNE embeddings on MNIST dataset for the active learning rounds 1, 5, 10, 15 and 20 using entropy sampling, our proposed MSAL strategy, and  the diversity-based MSAL-D strategies in Figure~\ref{fig:al-vis}.

For entropy sampling (Figure~\ref{subfig:tsne-entropy}), we observe that the embeddings change more dynamically across active learning rounds. In contrast, in our proposed methods (Figures~\ref{subfig:tsne-msal} and \ref{subfig:tsne-msald}), show relatively stable embeddings for each class. This stability supports our hypothesis that maintaining hyperspherical uniformity can be critical for improving representations in active learning. An interesting finding is that initial embeddings in Round 1 for our methods exhibit better separation with distinct clusters. Within our methods, MSAL-D selects more diverse samples for labeling from the unlabeled set compared to MSAL, which focuses more on uncertainty and sometimes fails to sample diverse data points in certain rounds. Furthermore, in Round 20, entropy sampling (\ref{subfig:tsne-entropy}) shows signs of feature degradation, where clusters become less discernible sometimes during the training. In contrast, our clusters formed around class prototypes become more compact over time, indicating improved feature representation. This comparative analysis highlights the effectiveness of our proposed MSAL and MSAL-D strategies over traditional uncertainty sampling in maintaining stable and informative embeddings throughout the active learning process.

\begin{figure}[ht]
    \centering
    \includegraphics[width=0.6\linewidth]{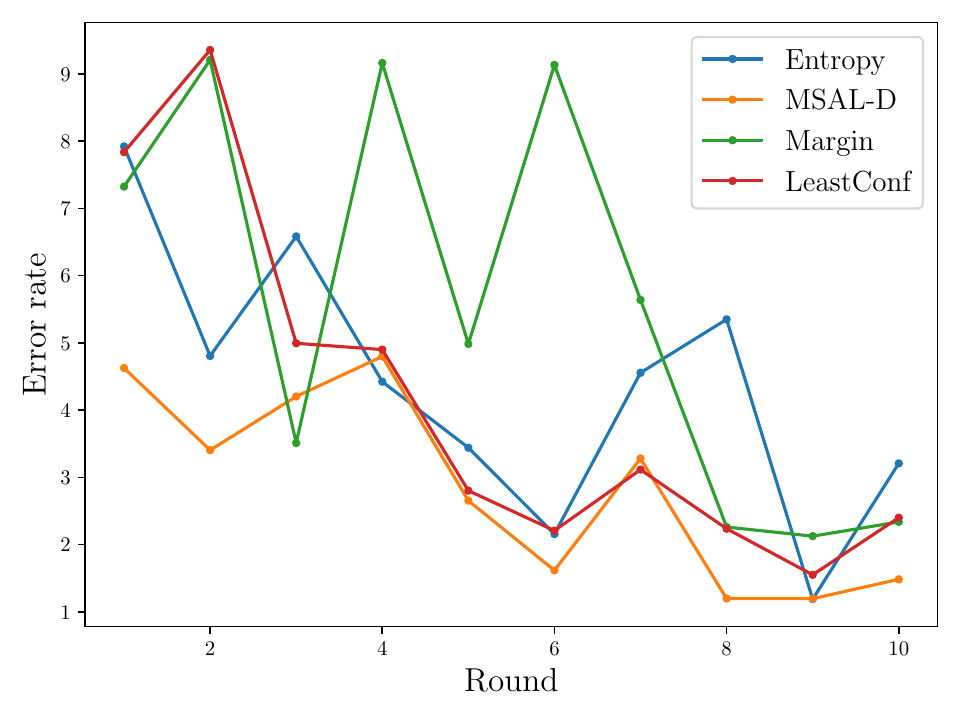}
    \caption{\textbf{Percentage of misclassified unlabeled samples} averaged across runs on MNIST dataset for entropy vs ours.}
    \label{fig:error_rate}
\end{figure}

\textbf{Error rate on unlabeled samples.} An active learning model is considered robust when it misclassifies as few data points as possible. We report the error rate of our method at every active learning round on MNIST dataset for the first 10 iterations. The error rate is the percentage of incorrect predictions on the total unlabeled data. A lower error rate at earlier stages indicates better initial representations. In Figure~\ref{fig:error_rate}, we show the for our proposed method, the misclassification rate is consistently lower than all other uncertainty-based strategies.

\textbf{Runtime comparison per dataset.} Since our proposed method, MSAL-D, does not have any clustering step unlike other combined strategies, the run-time of our method is comparable to other active learning methods, requiring only $6-10\%$ more time than entropy sampling.

\subsection{Hyperparameters}

\begin{figure}[ht]
    \centering
      \includegraphics[width=0.35\linewidth]{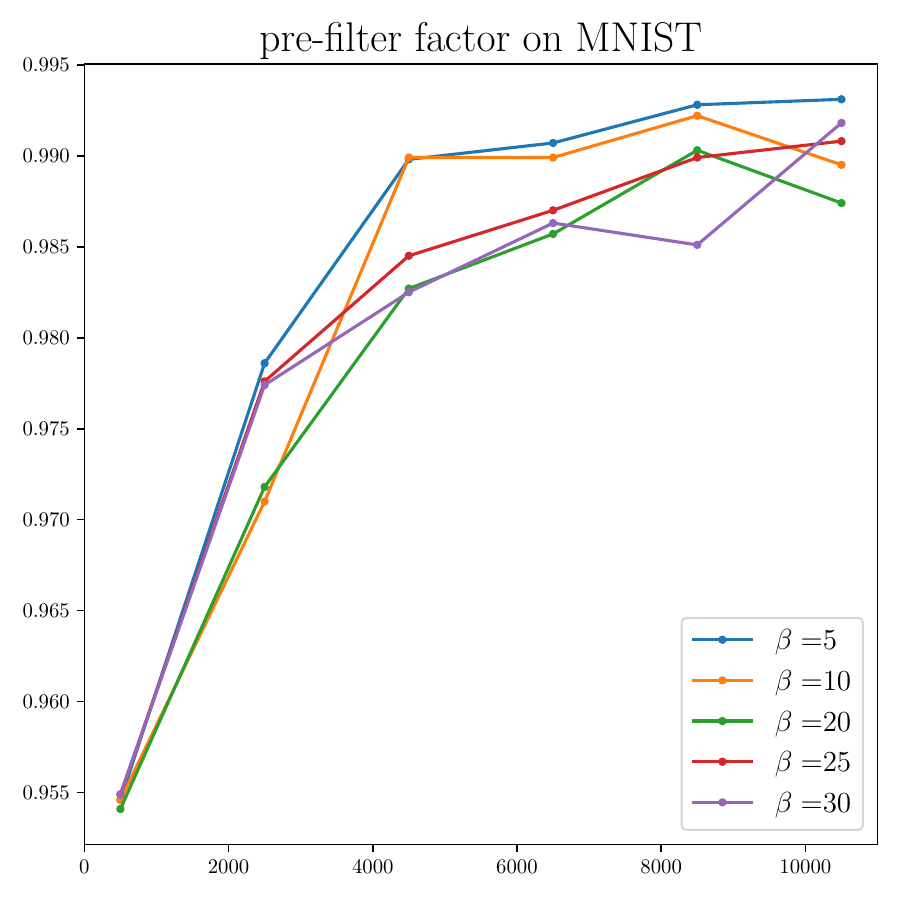}
    \includegraphics[width=0.35\linewidth]{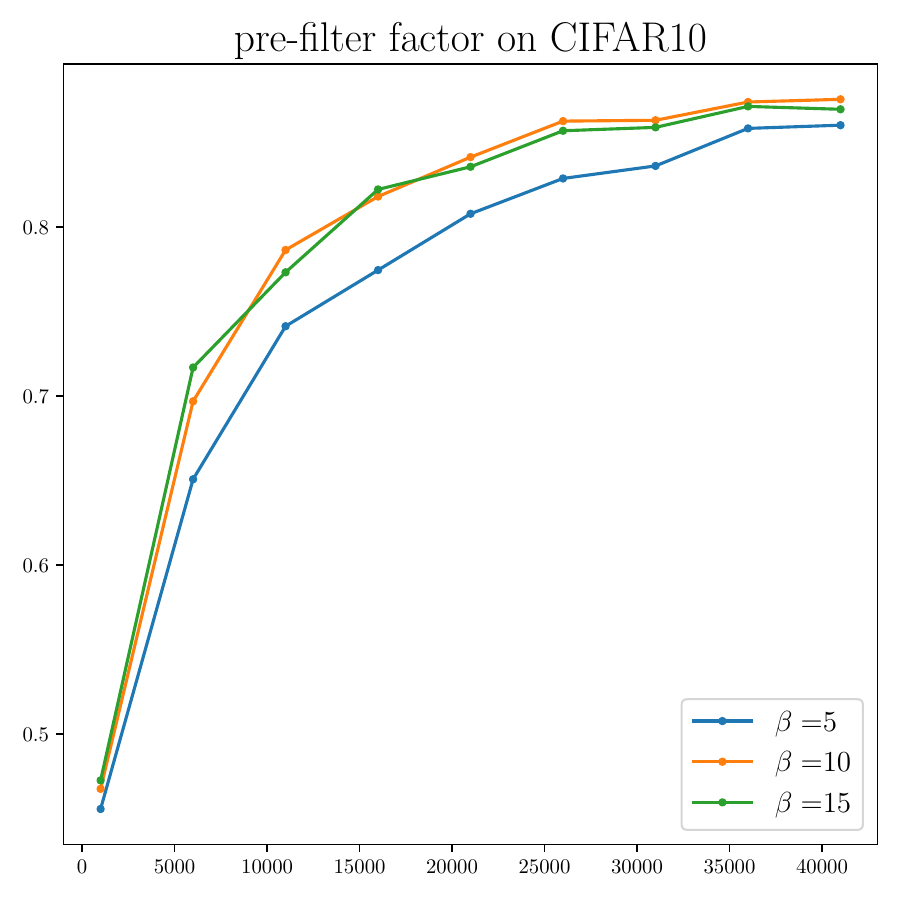}  

    \caption{\textbf{Overall accuracy vs. budget curves for $\mathbf{\beta}$} on \emph{MNIST} and \emph{SVHN} datasets.}
    \label{fig:beta-curves}
\end{figure}

\textbf{Pre-filter factor.} In our approach with diversity sampling, the pre-filter factor, $\beta$ decides how many uncertain samples $\beta b$ are considered for selecting the representative samples close to each class prototype. Higher the $\beta$ the samples closest to the class prototypes are considered, and also have lower uncertainty. We do a hyperparameter search for $\beta$ in $\{5,10,20,25,30\}$ for MNIST. We consider the overall performance using accuracy vs budget curve shown in Figure~\ref{fig:beta-curves}. We find that $\beta = 5$ performs the best for MNIST use the same hyperparameter for SVHN as the network configuration is the same. For CIFAR-10 we do a hyperparameter search in $\{5,10,15\}$, since lower values have shown to work better on prior datasets. We find that $\beta = 10$ is the best hyperparameter for CIFAR-10 and similarly use it for CIFAR-100 and TinyImagenet.

\textbf{Radius of hypersphere.} We use the heuristics as defined in the maximum separation work \cite{Kasarla2022} to choose the radius of the hypersphere. For MNIST, we use the radius $\rho = 0.1$ and for all other datasets we use $\rho = 1$. 

\begin{figure}{ht}
    \centering
    \includegraphics[width=0.9\linewidth]{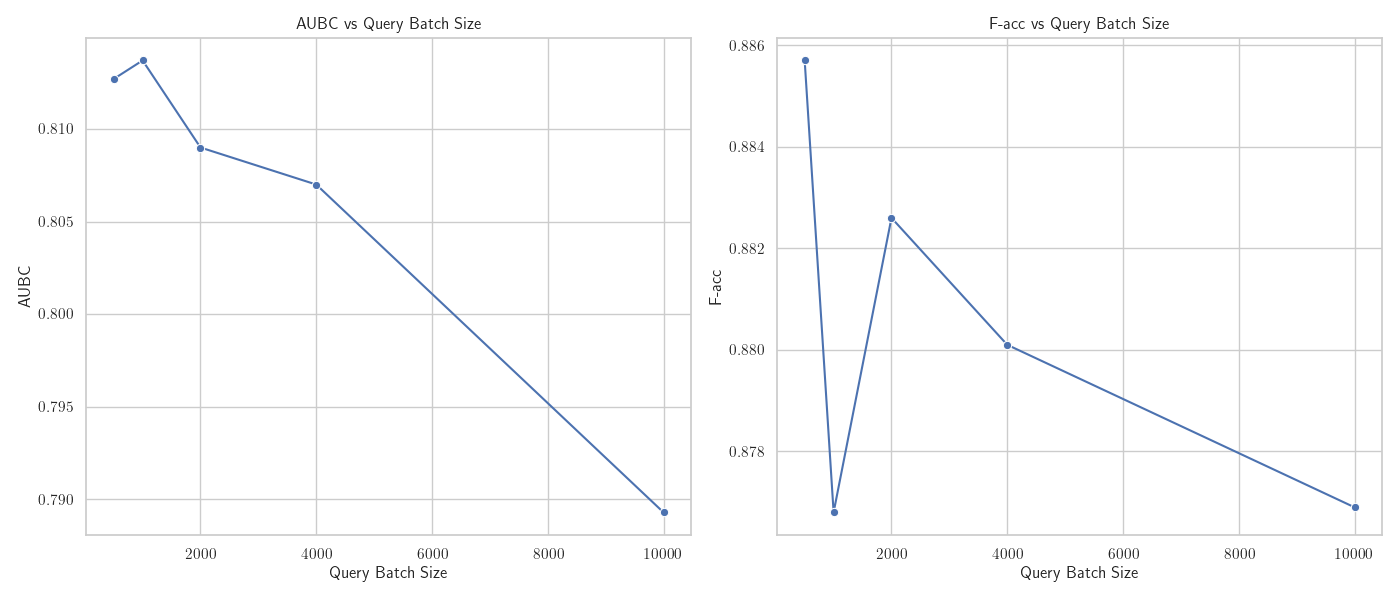}
    \caption{Ablation of AUBC and Accuracy(F-acc) vs query batch size per active learning round. We choose the batchsize with highest F-acc.}
    \label{fig:batch-size}
\end{figure}

\textbf{Effect of query size.} Here we conduct ablation studies on the effect of the query batch size $b$ on the overall performance of our proposed method on CIFAR-10 shown in Figure~\ref{fig:batch-size}. We notice that no value of batch size performs best on both AUBC and F-acc metrics, so we select the hyperparameter with the highest final accuracy, which is the batchsize $b=500$ used for all experiments CIFAR and TinyImageNet experiments.

\section{Conclusion}
\label{sec:conclusion}

This paper aims to integrate hyperspherical uniformity through maximum separation into active learning for image classification tasks. We propose an combined strategy that uses distance to fixed hyperspherical class prototypes to select both uncertain and representative samples for labeling. Additionally, the deep network embeds these fixed class prototypes at the end of final layer to learn maximally separable class representations. To demonstrate the effectiveness our proposal as a fundamental component for active learning methods, we perform various experiments on benchmark datasets. Our results show that (i) our proposed method outperforms other combined strategies (ii) multiple uncertainty-based strategies can be easily integrated with our method. (iii) our method is effective on imbalance AL datasets with potential for imbalanced specific methods in a future work. Overall, we conclude that enforcing maximum separation for active learning is key to learning robust representations from the start. 
\textbf{Limitations.} (a) While the method shows effectiveness on benchmark datasets, its performance on domain-specific or highly noisy datasets remains to be explored. (b) For the tasks that depends on the semantic similarity between classes, like zero-shot active learning~\cite{jia2021towards}, the assumption of equally separable classes, as in our case, may not be apt.

\textbf{Future Directions.} We currently show effectiveness of hyperspherical prototypes for AL on the problem of image classification. Such a technique may benefit other discriminative tasks like active learning for segmentation and question-answering as well, and provides an interesting future research direction.

\section*{Acknowledgements}

This work is financially supported by the ELLIS Amsterdam Unit. The compute used for this project has been partially supported by Vlaams Supercomputer Centrum and Dutch national supercomputer Snellius.

\bibliographystyle{splncs04}
\bibliography{main}
\end{document}